\documentclass[11pt,a4paper]{article}

\usepackage[T1]{fontenc}
\usepackage[utf8]{inputenc}
\usepackage{libertinus}         % Libertinus Serif + Sans + Math (modern serif stack)
\usepackage{microtype}

\usepackage[a4paper,top=2.6cm,bottom=2.8cm,left=2.9cm,right=2.9cm]{geometry}
\usepackage{setspace}
\usepackage{amsmath}
\usepackage{siunitx}
\usepackage{graphicx}
\usepackage{multirow}
\usepackage{array}
\usepackage[table,dvipsnames]{xcolor}
\usepackage{booktabs}
\usepackage{tabularx}
\usepackage{enumitem}
\usepackage{authblk}

\definecolor{pnavy}{HTML}{263238}      % blue-grey charcoal — primary "dark"
\definecolor{poxblood}{HTML}{C64545}   % crimson — bold accent
\definecolor{pochre}{HTML}{E9C46A}     % soft mustard yellow
\definecolor{pmoss}{HTML}{2A9D8F}      % persian green / turquoise
\definecolor{pplum}{HTML}{6D597A}      % muted violet — 4th distinct hue
\definecolor{pruleg}{HTML}{B0B0B0}     % neutral grey — rules
\definecolor{ppanel}{HTML}{FAF4E6}     % warm ivory — panels
\definecolor{plink}{HTML}{C64545}      % crimson for links (single accent)

\usepackage[explicit]{titlesec}
\titleformat{\section}
  {\normalfont\sffamily\Large\bfseries}
  {\thesection\hspace{0.75em}}{0em}{#1}
  [{\vspace{-4pt}\color{pruleg}\rule{\linewidth}{0.4pt}}]
\titleformat{\subsection}
  {\normalfont\sffamily\normalsize\bfseries}
  {\thesubsection\hspace{0.55em}}{0em}{#1}
\titleformat{\subsubsection}
  {\normalfont\sffamily\small\itshape}
  {\thesubsubsection\hspace{0.5em}}{0em}{#1}
\titlespacing*{\section}{0pt}{1.7em}{0.6em}
\titlespacing*{\subsection}{0pt}{1.2em}{0.35em}

\usepackage[font=small,labelfont={sf,bf},labelsep=period]{caption}

\arrayrulecolor{pruleg}

\newcommand{\rowH}{\rowcolor{gray!13}}

\usepackage[colorlinks=true,
            linkcolor=plink,
            citecolor=plink,
            urlcolor=plink,
            breaklinks=true]{hyperref}

\usepackage{tikz}
\usetikzlibrary{shapes.geometric,arrows.meta,positioning,fit,backgrounds,calc,matrix,decorations.pathreplacing}
\usepackage{pgfplots}
\pgfplotsset{compat=1.18}

\colorlet{aimsdark}{pnavy}
\colorlet{aimsaccent}{poxblood}
\colorlet{aimsrule}{pruleg}
\colorlet{aimspanel}{ppanel}
\colorlet{aimslink}{plink}

\usepackage[numbers,sort&compress,square,comma]{natbib}
\providecommand{\cortext}[2]{}
\providecommand{\corref}[1]{}

\providecommand{\journal}[1]{}

\begin{document}

\title{\Large\bfseries Detecting Money Laundering in Rwandan Mobile Money\\[3pt]
       \normalsize\mdseries\sffamily A Machine Learning Framework}

\author[1,$\ast$]{Emmanuel Nahimana}
\author[2,3]{Ya\'e Ulrich Gaba}
\affil[1]{African Institute for Mathematical Sciences (AIMS) Senegal, KM 2, Route de Joal, Mbour, Senegal}
\affil[2]{AI Research and Innovation Nexus for Africa (AIRINA Labs), AI.Technipreneurs, Cotonou, B\'enin}
\affil[3]{Sefako Makgatho Health Sciences University, Pretoria, South Africa}
\affil[$\ast$]{Corresponding author: \href{mailto:nahimana.emmanuel@aims-senegal.org}{\texttt{nahimana.emmanuel@aims-senegal.org}}}

\date{}

\maketitle

\noindent\hrulefill

\begin{abstract}
\noindent Mobile money has widened financial access across Sub-Saharan Africa and, in doing so, enlarged the surface for money-laundering and terrorism-financing (ML/TF) activity in ecosystems dominated by high-volume, low-value transactions. Rwanda is a case in point: several million active mobile-money users, telecom-led wallets on the MTN and Airtel networks, and a Financial Intelligence Centre (FIC) that must supervise transaction streams whose scale and velocity exceed the capacity of static rule-based monitoring. This paper develops and evaluates a transaction-monitoring framework aligned to the publicly documented constraints of the Rwandan AML/CFT regime under (i)~extreme class imbalance ($\sim 0.1\%$ laundering prevalence), (ii)~scarce and delayed labels, and (iii)~bounded investigator capacity. Using SAML-D, a synthetic transaction-monitoring dataset of $9{,}504{,}852$ transactions with $17$ laundering typologies, we engineer causal, account-centric behavioural features (rolling velocity, net-flow directionality, counterparty diversity, burstiness) and benchmark a portfolio of supervised classifiers (Logistic Regression, Random Forest, LightGBM), unsupervised anomaly detectors (Isolation Forest, Local Outlier Factor), a dense autoencoder, and a logistic-regression late-fusion meta-learner. Evaluation is operational rather than accuracy-based: PR-AUC, recall at a calibrated $\sim 90\%$-precision operating point, recall at top-$K\%$ of scored transactions, and alerts per $10{,}000$. On the chronologically held-out test period, LightGBM attains PR-AUC $=0.0469$ and captures $64$ true laundering cases at precision $\approx 0.89$ with $0.51$ alerts per $10{,}000$; the fusion stacker reaches PR-AUC $=0.0477$ at precision $\approx 0.91$ and $0.46$ alerts per $10{,}000$ while recovering $59$ true positives. Portfolio fusion therefore reduces alert volume marginally without adding recall in this regime. We map score bands to Rwanda-relevant analyst workflows and STR/SAR escalation, and outline a staged path from synthetic prototyping to secure real-data validation with the National Bank of Rwanda and FIC. The contribution is operational: a governance-aware pipeline and evaluation protocol calibrated to the constraints of an African mobile-money regulator, not a new algorithm.


\end{abstract}

\noindent\hrulefill

\vspace{0.6em}
\begin{center}
\begin{minipage}{0.94\linewidth}
{\footnotesize\sffamily\bfseries Highlights.}
\begin{itemize}[leftmargin=1.4em,itemsep=1pt,topsep=2pt,label=\raisebox{0.25ex}{\scriptsize\textcolor{poxblood}{\textbullet}}]\footnotesize
\item An AML transaction-monitoring pipeline aligned to Rwanda's FIC/BNR regime is developed and evaluated on SAML-D under 0.1\% laundering prevalence.
\item Causal, account-centric behavioural features are engineered under strict leakage controls and a chronological train/validation/test split.
\item LightGBM attains PR-AUC $=0.0469$ and captures $64$ true laundering transactions at precision $\approx 0.89$ with $0.51$ alerts per $10{,}000$.
\item Late-fusion of supervised, anomaly, and reconstruction scores yields a small alert-volume reduction (72$\to$65 alerts) at a small true-positive cost (64$\to$59) and does not improve calibrated recall on this dataset.
\item Model scores are mapped to a four-band analyst workflow with an explicit path from synthetic prototyping to secure real-data validation with BNR and FIC.
\end{itemize}
\end{minipage}
\end{center}
\vspace{0.4em}

%======================================================================
\section{Introduction}\label{sec:intro}

Money laundering and terrorism financing (ML/TF) impose real costs on financial integrity and distort investment in developing economies, where illicit flows are commonly estimated at a non-trivial fraction of GDP \citep{unodc2011,fatf2023recommendations}. In Sub-Saharan Africa, the rollout of mobile money has widened financial inclusion \citep{findex2021,gsma2023sotir} and, at the same time, enlarged the attack surface for illicit activity through micro-transaction structuring, agent-network abuse, weak KYC in some segments, and cross-border transfers whose provenance is hard to trace \citep{lokanan2023,gsmaregindex2021}.

Rwanda is a good case for study. The National Bank of Rwanda (BNR) reports several million active mobile-money users; MTN Mobile Money and Airtel Money dominate the retail channel; and the Financial Intelligence Centre (FIC) supervises AML/CFT reporting under national anti-money-laundering legislation \citep{bnr2022,rwandalaw2019}. The ESAAMLG Mutual Evaluation of Rwanda \citep{esaamlg2023} identifies gaps in institutional coordination, cross-platform visibility, and analytical capacity, all of which ML-based monitoring can help address.

Rule-based monitoring remains the operational default because it is auditable and simple to explain to supervisors. But it is brittle in mobile-money settings: fixed thresholds generate high false-positive volumes, fail to capture temporal or network structures typical of structuring and layering schemes, and adapt slowly to evolving typologies \citep{fatf2021newtech}. Machine learning provides a complementary approach that learns nonlinear feature interactions from historical activity and prioritises alerts by risk score rather than by rule violation \citep{jurgovsky2018,carcillo2021}.

\subsection*{Problem statement}
Operational AML in mobile-money-dominated ecosystems faces three interlocking constraints. (i)~\emph{Massive volume}: daily transaction counts overwhelm manual review; alert budgets are the binding constraint on any monitoring system. (ii)~\emph{Scarce and delayed labels}: confirmed suspicious-transaction reports (STRs) are a small fraction of activity and arrive months after the event, constraining supervised learning. (iii)~\emph{Adversarial adaptation}: typologies and user behaviour co-evolve with controls, degrading static thresholds and poorly monitored ML models. These are compounded by severe class imbalance: true laundering events represent approximately one in a thousand transactions in SAML-D, and comparable estimates hold in the literature \citep{bhattacharyya2011}.

\subsection*{Contributions}
This paper makes four contributions. First, we present a Rwanda-aligned monitoring pipeline: an end-to-end framework designed against the operational and regulatory constraints of the Rwandan AML/CFT regime (BNR, FIC, national AML legislation, ESAAMLG evaluation) rather than a generic benchmarking exercise on a synthetic dataset. Second, on top of this pipeline we engineer a family of causal, account-centric behavioural features (velocity, rolling aggregates, net flows, burstiness, counterparty diversity) computed strictly from past events and evaluated on a chronologically split test period to eliminate look-ahead leakage. Third, we conduct a portfolio evaluation using operational rather than accuracy-oriented metrics, namely PR-AUC, recall at a calibrated high-precision operating point, recall at top-$K\%$, alert budget per $10{,}000$ transactions, and confusion-matrix counts at the calibrated threshold; these are applied uniformly across supervised classifiers, anomaly detectors, a reconstruction-based autoencoder, and a late-fusion meta-learner. Finally, we translate model scores into a four-band risk-score schema linked to analyst-review actions and STR/SAR escalation, and outline a staged path from synthetic-data prototyping to secure real-data validation with the BNR and FIC.

We are explicit about what this paper is \emph{not}: it is not a new algorithm and does not claim generalisation beyond the SAML-D generative process. Section~\ref{sec:limits} states these limits directly.

\subsection*{Positioning against related work}
Our closest neighbours are (a) the SAML-D dataset paper of \citet{oztas2024saml}, which introduces the data and provides baseline experiments; (b) the mobile-payment fraud study of \citet{lokanan2023} for Sub-Saharan Africa; and (c) the hybrid supervised--unsupervised pipelines pioneered for card fraud \citep{carcillo2021}. Relative to \citet{oztas2024saml}, our contribution is the calibrated high-precision operating regime and the operational workflow mapping rather than algorithmic novelty; relative to \citet{lokanan2023}, we add causal temporal features, an explicit alert-budget formulation, and Rwanda-specific regulatory alignment; relative to \citet{carcillo2021}, we adapt the portfolio idea to mobile money and report that fusion mainly reduces alert volume without adding recall (Section~\ref{sec:results}).

Section~\ref{sec:regcontext} details the Rwandan regulatory and operational context. Section~\ref{sec:related} reviews related work. Section~\ref{sec:data} describes SAML-D and causal feature engineering. Section~\ref{sec:methods} presents models and fusion. Section~\ref{sec:eval} defines evaluation under severe imbalance and alert budgets. Section~\ref{sec:results} reports results, including confusion matrices, ablation, and recall at top-$K\%$. Section~\ref{sec:opmapping} maps scores to Rwanda-style investigation workflows. Section~\ref{sec:limits} discusses limitations, deployment considerations, and threats to validity, and Section~\ref{sec:conclusion} concludes.

%======================================================================
\section{Regulatory and operational context for Rwanda-style mobile money}\label{sec:regcontext}

An AML/CFT monitoring system in Rwanda operates within a specific institutional and legal architecture, and its usefulness is bounded by that architecture.

\subsection{Institutions and reporting flow}
Rwanda's AML/CFT regime rests on three pillars. National AML legislation defines predicate offences, reporting obligations, and sanctions \citep{rwandalaw2019}. The Financial Intelligence Centre (FIC), established under this law, receives STRs from reporting entities and disseminates intelligence to law enforcement. The National Bank of Rwanda supervises banks, payment institutions, and mobile-money issuers, and enforces prudential and AML compliance obligations \citep{bnr2022}. The ESAAMLG Mutual Evaluation Report on Rwanda \citep{esaamlg2023} sets out compliance gaps and drives national action plans.

At the operational level a mobile-money AML pipeline comprises four stages: (1)~transaction capture from mobile-money platforms (typically MTN Mobile Money and Airtel Money for the retail channel, and bank-led wallets for the higher-tier segment); (2)~internal monitoring and alert triage at the provider under supervisory scrutiny; (3)~case escalation to compliance and STR/SAR filing where thresholds are met; (4)~intelligence dissemination through FIC to competent authorities.

\subsection{Design constraints that dominate the problem}
Three constraints shape any monitoring design worth deploying.

\textit{Capacity is the binding constraint.} Investigation teams can review only a small fraction of daily activity, so alert volume is a first-order design variable rather than a secondary metric. A monitoring system that generates alerts at a rate exceeding investigator capacity cannot be operated, regardless of its detection metrics.

\textit{Explainability is required for supervision and adverse action.} Alerts must be defensible in observable behavioural terms (for instance, ``the account exhibited a $6\times$ increase in one-hour transaction velocity followed by fan-out to seven new counterparties within thirty minutes'') to support case narratives, STR filings, and any adverse action against a customer account. Scores that cannot be so decomposed create downstream cost for analysts and potential legal exposure.

\textit{Drift is continuous, not exceptional.} Product changes, seasonal patterns (agricultural cycles, remittance seasons, Umuganda and public holidays), and adversarial adaptation cause the distribution of activity to shift. Static thresholds silently degrade; ML models must be recalibrated on a schedule that supervisors can audit.

These constraints motivate a portfolio design: a supervised model for stable ranking, complementary anomaly detectors for rare-pattern discovery under strict alert budgets, and a conservative fusion layer for score corroboration and explainability rather than for recall gains.

\subsection{What is Rwanda-specific about mobile-money laundering typologies}
Prior work in comparable Sub-Saharan African markets \citep{lokanan2023,gsmaregindex2021} identifies typologies of particular concern in mobile-money-dominated ecosystems: (a)~structuring through repeated cash-in operations near KYC-tier thresholds; (b)~fan-in patterns where multiple small transfers concentrate on a merchant or business wallet; (c)~agent-mediated cash conversion that mimics structuring at the sub-hour scale; (d)~cross-border micro-transfers exploiting inconsistent KYC across neighbouring jurisdictions; (e)~dormant-wallet reactivation with atypical volume. Our feature engineering (Section~\ref{sec:data}) is designed to encode signals for each of these typology families.

%======================================================================
\section{Related work}\label{sec:related}

Having situated the Rwandan operating environment, we now review the technical literature that informs the classifier portfolio and the evaluation choices in the sections that follow.

\subsection{From rule-based monitoring to machine learning}
Historically, AML transaction monitoring has relied on rule-based systems: static thresholds on amount, frequency, and counterparty attributes, combined with typology-driven scenarios (structuring near reporting thresholds, rapid movement across accounts, layering through third parties). Such systems are auditable and easily explained to supervisors, but their false-positive rates are frequently reported to exceed 90--95\% in production, saturating investigation queues and displacing analyst attention from genuinely suspicious activity \citep{fatf2021newtech}. Machine learning entered the AML stack first as an alert-prioritisation layer over existing rules, then as a stand-alone ranking system in mature deployments. Our framework belongs to the second category: we replace the rule-based scorer with a supervised model and preserve the rule-based signal only as an audit anchor.

\subsection{Supervised, anomaly, and hybrid pipelines}
Supervised learning remains the workhorse of AML alerting when labels are available. Linear baselines (Logistic Regression) offer interpretability that supervisors accept; tree ensembles (Random Forest, gradient boosting) capture nonlinear interactions between amount, frequency, and counterparty structure \citep{bhattacharyya2011,ke2017lightgbm}. Anomaly detectors (Isolation Forest \citep{liu2008iforest}, Local Outlier Factor \citep{breunig2000lof}) and reconstruction-based autoencoders \citep{chalapathy2019} contribute complementary signals when labels are scarce or delayed. \citet{carcillo2021} demonstrate the benefit of combining supervised and unsupervised signals in card fraud; the design generalises to any high-imbalance transactional setting where labels are late and typologies are non-stationary. Our portfolio adapts that idea to mobile money and reports that, on SAML-D, the fusion benefit is modest and non-monotone in the operating point (Section~\ref{sec:fusiondiscussion}).

\subsection{Graph, sequence, and attention-based approaches}
Beyond tabular pipelines, graph-based methods use transaction networks to detect multi-hop laundering structures. \citet{weber2019bitcoin} apply graph convolutional networks to a labelled subgraph of the Bitcoin blockchain (the ``Elliptic'' dataset), demonstrating that account-level graph features materially improve detection of layering patterns that are invisible to per-transaction classifiers. Sequence models capture long-range temporal dependencies in an account's own activity: attention-based architectures adapted from natural-language processing have been used to encode transactional histories, and recurrent architectures (LSTM, GRU) provide simpler alternatives for shorter horizons \citep{jurgovsky2018}. These directions are promising but computationally demanding and, in production settings, harder to defend to supervisors than tree ensembles with SHAP-based post-hoc explanations. In an African mobile-money deployment they face an additional constraint: account-level linkage across providers (MTN, Airtel, bank-led wallets) is not currently technically or legally feasible at the granularity required to build a useful cross-provider graph. We treat both graph and long-context sequence approaches as complementary future work, activated once the underlying data architecture and legal framework permit them.

\subsection{Privacy, federation, and cross-institution collaboration}
A recurring theme in the recent AML literature is the tension between the value of pooling data across institutions (banks, mobile-money providers, regulators) and the legal and reputational cost of moving raw transaction data across institutional boundaries. Federated learning has been proposed as a middle path: models are trained collaboratively across data holders while the raw data remains local. This paradigm is attractive for Rwanda's fragmented mobile-money landscape (multiple providers, bank-led wallets, a supervisory perimeter that spans BNR and FIC), but it presupposes a shared feature space, coordinated schema versioning, and secure aggregation infrastructure that few African regulators currently operate. We flag federation as a directional recommendation rather than an operational option for this work.

\subsection{Reproducibility, synthetic benchmarks, and the evaluation gap}\label{sec:relwork-synthetic}
Public real-data AML benchmarks are scarce because transaction data is confidential and the ground-truth (confirmed STR outcomes) is subject to disclosure restrictions. This has produced a well-documented reproducibility gap in the field: many claimed improvements are reported on proprietary or lightly documented internal data, and cross-paper comparison is often impossible. Synthetic simulators fill part of this gap. PaySim \citep{lopezrojas2016paysim} was the first widely-adopted mobile-money fraud simulator and remains a common reference. The IBM AML-World generator \citep{altman2023amlworld} produces multi-typology transaction graphs at large scale and has been positioned as a shared benchmark for graph-based methods. SAML-D \citep{oztas2024saml} focuses on transaction-level monitoring with 17 explicit laundering typologies and is the dataset we adopt. Synthetic data enables reproducible benchmarking and stress-testing under controlled typology mixes, but it cannot replicate the full distribution of operational noise (missingness, schema drift, delayed corrections), label uncertainty (mis-filed STRs, provisional versus confirmed outcomes), or adversarial adaptation. Any result reported on a synthetic benchmark, including ours, should be read as evidence for a pipeline and evaluation protocol, not as a production-calibrated performance number.

\subsection{African and mobile-money--specific evidence}
The mobile-money vulnerabilities most cited in the Sub-Saharan African context concern SIM-swap-related account takeover, agent collusion (especially at high-turnover agents in border regions), weak or inconsistently applied KYC in low-tier customer segments, and cross-border micro-transfers between neighbouring jurisdictions with divergent AML regimes \citep{gsmaregindex2021,gsma2023sotir,lokanan2023}. These vulnerabilities are structural rather than algorithmic: they cannot be resolved by a better classifier alone, but a well-designed classifier makes the residual analytical burden tractable. In Rwanda, the ESAAMLG Mutual Evaluation \citep{esaamlg2023} identifies analytical capacity and cross-provider coordination as specific gaps, both of which a portfolio-based ML monitoring system is intended to help close. GSMA's regulatory indices \citep{gsmaregindex2021,gsma2023sotir} identify the same set of vulnerabilities in cross-country comparative form, and motivate the behavioural feature families (velocity, flow directionality, counterparty diversity) that we adopt in Section~\ref{sec:data}.

%======================================================================
\section{Data and feature engineering}\label{sec:data}

\subsection{Dataset and chronological splits}
We use SAML-D \citep{oztas2024saml,oztaskaggle2023}, a synthetic dataset for AML transaction monitoring with $9{,}504{,}852$ transactions after basic cleaning. Each transaction carries a binary label \texttt{Is\_laundering} and a categorical \texttt{Laundering\_type} taking one of 17 typology values on positive cases. Laundering prevalence is approximately $0.1\%$, mirroring the imbalance observed in real AML systems.

To emulate deployment and avoid look-ahead bias, the dataset is split \emph{chronologically} into three disjoint subsets by timestamp: a training set of $6.65$M transactions (early period), a validation set of $1.43$M transactions (middle period, used for hyperparameter tuning and threshold calibration), and a test set of $1.43$M transactions (most recent period, used exclusively for final evaluation). All engineered features are computed causally: for a transaction at time $t_i$, features aggregate only events with timestamps $t_j < t_i$ for the same account. The \texttt{Laundering\_type} field and any label-derived variables are excluded from model inputs to prevent leakage.

\subsection{Causal behavioural features}
Let $\mathcal{W}_a(\Delta t)$ denote the set of transactions of account $a$ within the time window of length $\Delta t$ ending strictly before the scored event, and let $A_i$ denote transaction amounts. We construct the following feature groups.

\textit{Rolling activity} — transaction counts and summed amounts over windows of $1$~hour, $24$~hours, and $7$~days, targeting structuring and bursty behaviour.

\textit{Velocity} — normalised activity intensity,
\begin{equation}
V_a(\Delta t) \;=\; \frac{\sum_{i \in \mathcal{W}_a(\Delta t)} A_i}{\Delta t},
\end{equation}
computed at multiple horizons.

\textit{Inter-arrival times and burstiness} — for successive transactions of the same account, $\mathrm{IAT}_i = t_i - t_{i-1}$; and a bounded burstiness index following the standard construction \citep{goh2008burstiness},
\begin{equation}
B \;=\; \frac{\sigma(\Delta t) - \mu(\Delta t)}{\sigma(\Delta t) + \mu(\Delta t)} \;\in\; (-1, 1),
\end{equation}
distinguishing regular from bursty tempo.

\textit{Flow directionality} — signed amounts, net inflow/outflow, and a cash-in vs cash-out asymmetry index
\begin{equation}
\alpha_i \;=\; \frac{\mathrm{cash\_in}_i - \mathrm{cash\_out}_i}{\mathrm{cash\_in}_i + \mathrm{cash\_out}_i + \varepsilon},
\end{equation}
proxying mule and pass-through behaviour.

\textit{Counterparty diversity} — the number of distinct counterparties over a window, and a Herfindahl-style concentration index
\begin{equation}
H_a(\Delta t) \;=\; \sum_{c \in \mathcal{C}_a(\Delta t)} p(c)^2,
\end{equation}
targeting fan-in / fan-out and repeated-ring structures.

\textit{Temporal indicators} — hour-of-day, day-of-week, and flags for night-time and weekend activity to capture off-hours anomalies relative to baseline.

Table~\ref{tab:features} summarises the feature groups and their AML motivation.

\begin{table}[t]
\centering
\caption{Feature groups used for risk scoring (causal, account-centric).}\label{tab:features}
\small
\begin{tabular}{p{3.2cm}p{4.2cm}p{6.2cm}}
\toprule
\rowH \textbf{Group} & \textbf{Examples (illustrative)} & \textbf{AML motivation}\\
\midrule
Rolling activity & \texttt{cnt\_1h}, \texttt{sum\_24h}, \texttt{sum\_7d} & Structuring; bursty micro-transaction sequences.\\
Velocity \& tempo & \texttt{velocity\_1h}, \texttt{iat}, \texttt{burstiness} & Rapid layering; automation-like tempo.\\
Net flows & \texttt{net\_24h}, \texttt{in\_out\_ratio} & Fan-in / fan-out; cycling and cash-in-cash-out loops.\\
Counterparty diversity & \texttt{uniq\_cp\_24h}, \texttt{H\_7d} & Rings; mule dispersion and hub concentration.\\
Time context & \texttt{hour}, \texttt{dow}, \texttt{is\_night} & Off-hours anomalies relative to account baseline.\\
Amount shape & \texttt{log\_amt}, \texttt{amt\_zscore} & Threshold gaming; unusual amount distributions.\\
\bottomrule
\end{tabular}
\end{table}

\subsection{Preprocessing and imbalance handling}
Continuous features are standardised on the training set only and applied unchanged to validation and test; categorical variables (channel, transaction type) are one-hot encoded with training-set vocabulary. Missing values in SAML-D are structurally absent by construction; in a real deployment, imputation would be fit on the training window only.

To mitigate imbalance we use \emph{cost-sensitive learning} (\texttt{class\_weight="balanced"} for Logistic Regression and Random Forest; \texttt{scale\_pos\_weight} calibrated for LightGBM) rather than SMOTE-style oversampling. Synthetic minority resampling would break the temporal structure of a chronological split and inject near-duplicates that inflate optimistic test-set performance. Sensitivity experiments varying the class-weight parameter and negative-to-positive sampling ratio showed PR-AUC and recall at high precision stable within $\pm 5\%$ across configurations.

%======================================================================
\section{Methods}\label{sec:methods}

\begin{figure}[t]
\centering
\begin{tikzpicture}[
  font=\footnotesize,
  node distance=4mm and 5mm,
  every node/.style={align=center},
  data/.style={rectangle, rounded corners=2pt, draw=aimsdark, thick, fill=aimspanel, minimum width=22mm, minimum height=9mm},
  proc/.style={rectangle, rounded corners=2pt, draw=aimsdark, thick, fill=aimsdark!12, minimum width=22mm, minimum height=9mm},
  model/.style={rectangle, rounded corners=2pt, draw=aimsdark, thick, fill=white, minimum width=17mm, minimum height=7mm},
  outbox/.style={rectangle, rounded corners=2pt, draw=aimsaccent, thick, fill=aimsaccent!15, minimum width=22mm, minimum height=9mm},
  arr/.style={-Latex, thick, aimsdark!70}
]
% row 1
\node[data] (raw) {Raw SAML-D\\ \tiny 9.5M transactions};
\node[proc, right=of raw] (split) {Chronological\\ split (70/15/15)};
\node[proc, right=of split] (feat) {Causal feature\\ engineering};

% row 2 — models
\node[below=10mm of feat, xshift=-18mm] (label) {\footnotesize\bfseries Portfolio scoring};
\node[model, below=3mm of label, xshift=-24mm] (sup) {LightGBM,\\ RF, LR};
\node[model, right=3mm of sup] (unsup) {IF, LOF};
\node[model, right=3mm of unsup] (ae) {Autoencoder};

% row 3
\node[proc, below=6mm of unsup] (fuse) {Late fusion\\ (logistic stacker)};
\node[outbox, right=of fuse] (band) {Risk banding\\ A / B / C / D};
\node[outbox, right=of band] (act) {Analyst\\ workflow \& STR};

\draw[arr] (raw) -- (split);
\draw[arr] (split) -- (feat);
\draw[arr] (feat.south) -- (sup.north);
\draw[arr] (feat.south) -- (unsup.north);
\draw[arr] (feat.south) -- (ae.north);
\draw[arr] (sup) -- (fuse);
\draw[arr] (unsup) -- (fuse);
\draw[arr] (ae) -- (fuse);
\draw[arr] (fuse) -- (band);
\draw[arr] (band) -- (act);
\end{tikzpicture}
\caption{End-to-end AML monitoring pipeline: raw transaction stream, chronological train/validation/test split, causal account-centric feature engineering, portfolio scoring across supervised and anomaly detectors, logistic late-fusion, and threshold-calibrated risk banding feeding analyst workflow and STR/SAR reporting.}
\label{fig:pipeline}
\end{figure}
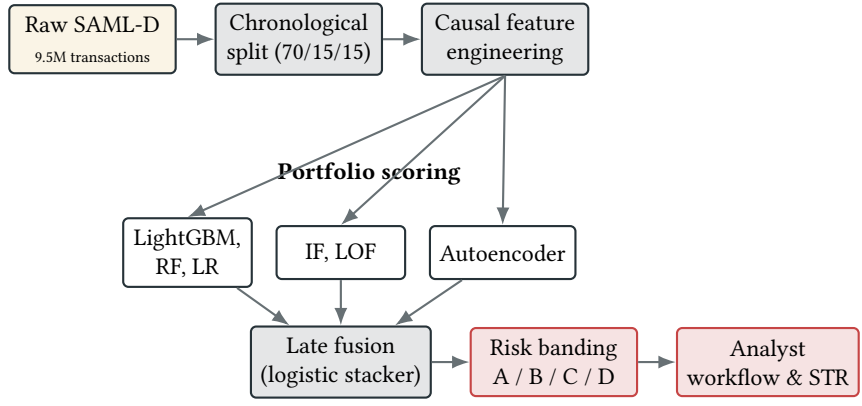

Figure~\ref{fig:pipeline} summarises the end-to-end monitoring pipeline: chronological ingestion and split, causal feature engineering, portfolio scoring across supervised and anomaly detectors, late-fusion aggregation, and threshold-calibrated banding for analyst review. Each stage is described in turn below.

\subsection{Supervised models}
We benchmark three supervised classifiers.

\textit{Logistic Regression} — $\ell_2$-regularised ($C=1.0$), \texttt{saga} solver, $\mathtt{class\_weight}=\mathtt{balanced}$, max $200$ iterations, on standardised features. Serves both as an interpretable baseline and as the meta-learner for late fusion.

\textit{Random Forest} \citep{breiman2001} — $T=500$ trees, unlimited depth subject to \texttt{min\_samples\_leaf} constraints, \texttt{balanced\_subsample} class weighting; hyperparameters (max depth, min samples per leaf) tuned by PR-AUC on the validation window.

\textit{LightGBM} \citep{ke2017lightgbm} — histogram-based leaf-wise gradient boosting; learning rate $0.02$--$0.05$; up to $800$ boosting iterations with early stopping on validation PR-AUC; \texttt{num\_leaves}~$=64$; \texttt{min\_data\_in\_leaf}~$=200$; \texttt{feature\_fraction}~$=0.8$; \texttt{scale\_pos\_weight} calibrated to the empirical class ratio. Treated as the primary supervised detector.

Hyperparameter selection for both Random Forest and LightGBM used a small-grid random search ($40$ configurations per model) with PR-AUC on the validation window as the selection metric; the best-validation configuration was then evaluated once on the held-out test window.

\subsection{Unsupervised and reconstruction detectors}
\textit{Isolation Forest} \citep{liu2008iforest} — $300$ trees, subsample size $10{,}000$; contamination parameter tuned to produce a realistic alert rate on the validation window; anomaly scores rescaled to $[0,1]$.

\textit{Local Outlier Factor} \citep{breunig2000lof} — $k=50$ neighbours, Euclidean metric; run in \texttt{novelty} mode with the reference set drawn from the training window's non-laundering transactions.

\textit{Autoencoder} — dense feed-forward, encoder $[128, 32, 8]$ with a $d=8$ latent code, decoder $[8, 32, 128]$, ReLU activations, Adam optimiser with initial learning rate $10^{-3}$, batch size $1024$, MSE reconstruction loss, early stopping (patience $10$) on validation reconstruction loss. Trained on non-laundering transactions from the training window only. The reconstruction error $\|x - \hat{x}\|^2$ is used as anomaly score, thresholded via the $99.5$th percentile on non-laundering validation transactions.

The autoencoder architecture is shown in Figure~\ref{fig:ae}. Sequence models (LSTM, GRU) were considered as an optional re-scoring layer but did not add calibrated recall over LightGBM on this dataset in preparatory experiments; they are noted here for the deployment path in Section~\ref{sec:limits} rather than as a reported result.

\begin{figure}[t]
\centering
\begin{tikzpicture}[
  font=\footnotesize,
  layer/.style={rectangle, rounded corners=1.5pt, draw=aimsdark, thick, minimum width=14mm, minimum height=8mm, align=center},
  encoder/.style={layer, fill=aimsdark!15},
  latent/.style={layer, fill=aimsaccent!25, minimum width=10mm},
  decoder/.style={layer, fill=aimsdark!8},
  arr/.style={-Latex, thick, aimsdark!75},
  node distance=2mm and 4mm
]
\node[encoder] (in) {input\\ \tiny $d_{\text{in}}$};
\node[encoder, right=of in] (e1) {dense\\ \tiny 128};
\node[encoder, right=of e1] (e2) {dense\\ \tiny 32};
\node[latent, right=of e2] (lat) {latent\\ \tiny $d{=}8$};
\node[decoder, right=of lat] (d1) {dense\\ \tiny 32};
\node[decoder, right=of d1] (d2) {dense\\ \tiny 128};
\node[decoder, right=of d2] (out) {$\hat{x}$\\ \tiny $d_{\text{in}}$};

\draw[arr] (in) -- (e1);
\draw[arr] (e1) -- (e2);
\draw[arr] (e2) -- (lat);
\draw[arr] (lat) -- (d1);
\draw[arr] (d1) -- (d2);
\draw[arr] (d2) -- (out);

\draw[decorate,decoration={brace,amplitude=4pt,mirror}]
  ($(in.south west)+(0,-2mm)$) -- ($(e2.south east)+(0,-2mm)$)
  node[midway,below=3pt,font=\scriptsize,color=aimsdark]{Encoder};
\draw[decorate,decoration={brace,amplitude=4pt,mirror}]
  ($(d1.south west)+(0,-2mm)$) -- ($(out.south east)+(0,-2mm)$)
  node[midway,below=3pt,font=\scriptsize,color=aimsdark]{Decoder};

\node[below=10mm of lat, font=\scriptsize, align=center] {%
Anomaly score $=\|x-\hat{x}\|^2$;\\
threshold at $99.5$th pct.\ of validation non-laundering.};
\end{tikzpicture}
\caption{Dense autoencoder used as the reconstruction-based anomaly detector: symmetric encoder--decoder $[128,32,8,32,128]$ with a $d=8$ latent bottleneck, trained on non-laundering transactions only, thresholded on the $99.5$th percentile of reconstruction error on validation.}
\label{fig:ae}
\end{figure}

\subsection{Late fusion via logistic stacking}
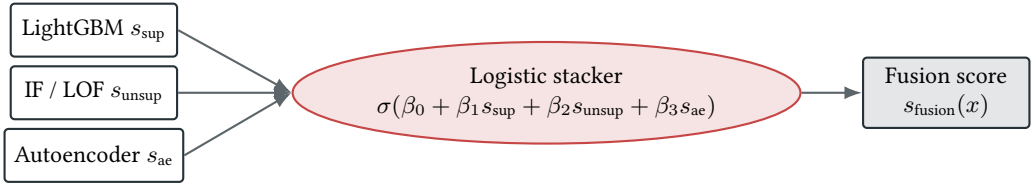
\begin{figure}[t]
\centering
\begin{tikzpicture}[
  font=\footnotesize,
  every node/.style={align=center},
  base/.style={rectangle, rounded corners=1.5pt, draw=aimsdark, thick, minimum width=22mm, minimum height=7mm, fill=white},
  meta/.style={ellipse, draw=aimsaccent, thick, fill=aimsaccent!15, minimum width=32mm, minimum height=10mm},
  outbox/.style={rectangle, rounded corners=1.5pt, draw=aimsdark, thick, minimum width=22mm, minimum height=7mm, fill=aimsdark!12},
  arr/.style={-Latex, thick, aimsdark!75}
]
\node[base] (lgbm) {LightGBM $s_{\text{sup}}$};
\node[base, below=1mm of lgbm] (iflof) {IF / LOF $s_{\text{unsup}}$};
\node[base, below=1mm of iflof] (ae) {Autoencoder $s_{\text{ae}}$};

\node[meta, right=15mm of iflof] (stack) {Logistic stacker\\ $\sigma(\beta_0+\beta_1 s_{\text{sup}}+\beta_2 s_{\text{unsup}}+\beta_3 s_{\text{ae}})$};
\node[outbox, right=8mm of stack] (score) {Fusion score\\ $s_{\text{fusion}}(x)$};

\draw[arr] (lgbm.east) -- (stack.west);
\draw[arr] (iflof.east) -- (stack.west);
\draw[arr] (ae.east) -- (stack.west);
\draw[arr] (stack.east) -- (score.west);
\end{tikzpicture}
\caption{Late-fusion (stacking) architecture: base-detector scores are fed into a logistic-regression meta-learner fitted on the validation window; coefficients $\beta_k$ are class-weighted. In practice on SAML-D the fitted stacker is dominated by $\beta_1$ (LightGBM), rendering the fusion score effectively a stricter cut of the supervised score at the high-precision operating point (see \S\ref{sec:fusiondiscussion}).}
\label{fig:fusion}
\end{figure}

Given the six base-model scores at each transaction --- from Logistic Regression, Random Forest, LightGBM (supervised), Isolation Forest, LOF (unsupervised), and the autoencoder --- the fusion score is
\begin{equation}
s_{\mathrm{fusion}}(x) \;=\; \sigma\!\left(\beta_0 + \sum_{k=1}^{6} \beta_k\, s_k(x)\right),
\label{eq:fusion}
\end{equation}
where $\sigma$ is the logistic function, $s_k$ ranges over the six base scores in the order above, and coefficients $\{\beta_k\}$ are fit on the validation window using the same class-weighting scheme. The stacker is illustrated in Figure~\ref{fig:fusion}. Fusion is designed as \emph{corroboration}, not as an independent alert source: at the calibrated high-precision operating point, unsupervised evidence lifts a transaction into the alert band only when the supervised score is already high. Section~\ref{sec:results} reports the empirical consequence of this design.

%======================================================================
\section{Evaluation under extreme imbalance and alert budgets}\label{sec:eval}

At laundering prevalence $\pi \approx 0.001$, ROC-AUC becomes an unreliable operational signal: a model with high ROC-AUC can still generate an unsupportable false-positive rate at any usable precision. We therefore evaluate on four metrics aligned with an AML operations team.

\textit{PR-AUC.} Precision--Recall area under the curve, more informative than ROC-AUC in the rare-positive regime \citep{davis2006prroc}. Reported as the primary ranking metric.

\textit{Recall at $\sim 90\%$ precision.} For each model we select on the validation window the smallest threshold $\tau$ such that $\mathrm{Precision}(\tau) \geq 0.90$, and report the resulting recall on the held-out test window. This is the natural operating regime for a supervisor-facing system where each alert triggers an analyst-hour of investigation.

\textit{Alerts per $10{,}000$ transactions.} A direct proxy for investigator workload at the chosen operating point.

\textit{Recall at top-$K\%$.} The fraction of true laundering transactions captured within the highest-scoring $K\%$ of the test period. This metric aligns evaluation with a capacity-based triage policy in which analysts work from the top of the ranked queue downward.

\subsection*{Threshold calibration}
Two complementary calibration rules are used.

\textit{High-precision operating point.} Select the smallest $\tau$ on the validation window such that $\mathrm{Prec}(\tau) \geq 0.90$, and among feasible thresholds choose the one maximising recall.

\textit{Budgeted alerting.} Select $\tau$ to satisfy an alert budget $B$ per $10{,}000$ transactions ($\mathrm{Alerts/10k}(\tau) \leq B$) and maximise recall subject to the constraint.

We report primarily the high-precision regime. The budgeted regime is used in the ablation and sensitivity analysis (Section~\ref{sec:ablation}).

%======================================================================
\section{Results}\label{sec:results}

\begin{center}
\fbox{\begin{minipage}{0.92\linewidth}\small
\textbf{Note on synthetic data.} All numbers reported in this section are computed on the SAML-D synthetic dataset. They do not represent BNR- or FIC-observable performance; they are evidence for the pipeline and evaluation protocol, and must be recalibrated on real transaction data before any deployment decision. See Section~\ref{sec:limits}.
\end{minipage}}
\end{center}

\subsection{Supervised and fusion performance at the high-precision operating point}
The headline result is that late-fusion does not improve calibrated recall over LightGBM on this dataset; it reduces alert volume marginally at a small true-positive cost. Table~\ref{tab:sup} reports test-set performance at the validation-calibrated $\sim 90\%$-precision operating point. LightGBM attains the strongest single-model PR-AUC ($0.0469$), captures $64$ true laundering cases at precision $0.889$, and emits $0.51$ alerts per $10{,}000$ transactions. The fusion stacker reaches marginally higher PR-AUC ($0.0477$) at slightly higher precision ($0.908$) and lower alert volume ($0.46$ per $10{,}000$), and recovers $59$ true positives (five fewer than LightGBM). This trade is discussed further in Section~\ref{sec:fusiondiscussion}.

\begin{figure}[t]
\centering
\begin{tikzpicture}
\begin{axis}[
  width=0.85\linewidth, height=6.2cm,
  xlabel={Recall},
  ylabel={Precision},
  xmin=0, xmax=0.4,
  ymin=0.0005, ymax=1.05,
  ymode=log,
  log basis y={10},
  ytick={0.001,0.01,0.1,1},
  yticklabels={$10^{-3}$,$10^{-2}$,$10^{-1}$,$10^{0}$},
  grid=both,
  minor grid style={line width=.1pt,draw=aimsrule!50},
  major grid style={line width=.15pt,draw=aimsrule},
  axis line style={aimsdark!70},
  tick label style={font=\footnotesize},
  label style={font=\footnotesize\color{aimsdark}},
  legend style={font=\scriptsize,at={(0.98,0.98)},anchor=north east,draw=aimsrule,fill=white,fill opacity=0.9,text opacity=1},
  legend cell align=left,
  every axis plot/.append style={mark size=1.6pt, thick}
]
% Random baseline (prevalence ~0.0012)
\addplot[dashed, gray!70, thick, mark=none] coordinates {(0,0.0012) (0.4,0.0012)};
\addlegendentry{Random ($\pi\!\approx\!0.0012$)}

% Logistic Regression — ochre triangles
\addplot[color=pochre, mark=triangle*, mark options={fill=pochre}] coordinates {
  (0.0035,1.000) (0.104,0.0124) (0.203,0.00806) (0.268,0.00639)
};
\addlegendentry{Logistic Regression (AP $=$ 0.012)}

% Random Forest — moss diamonds
\addplot[color=pmoss, mark=diamond*, mark options={fill=pmoss}] coordinates {
  (0.033,0.889) (0.080,0.00953) (0.141,0.00560) (0.207,0.00493)
};
\addlegendentry{Random Forest (AP $=$ 0.040)}

% LightGBM — deep navy, thick, filled circle (primary)
\addplot[color=pnavy, very thick, mark=*, mark options={fill=pnavy}] coordinates {
  (0.0377,0.889) (0.139,0.0166) (0.253,0.01004) (0.342,0.00814)
};
\addlegendentry{LightGBM (AP $=$ 0.047)}

% Fusion — oxblood, thick, dashed square
\addplot[color=poxblood, very thick, mark=square*, mark options={fill=poxblood}, dashed] coordinates {
  (0.0348,0.908) (0.126,0.0151) (0.246,0.00977) (0.341,0.00812)
};
\addlegendentry{Fusion stacker (AP $=$ 0.048)}
\end{axis}
\end{tikzpicture}
\caption{Precision--recall performance on the held-out test set for the four ranking models. Points are the validation-calibrated $\sim 90\%$-precision operating point (leftmost anchor per curve) and the top-$K\%$ ranking anchors for $K\in\{1,3,5\}$. Precision axis is logarithmic to make the high-precision regime legible. The dashed grey line is the random baseline at test-set prevalence $\pi\approx 0.0012$; every model sits at least an order of magnitude above it across the plotted recall range.}
\label{fig:prcurves}
\end{figure}
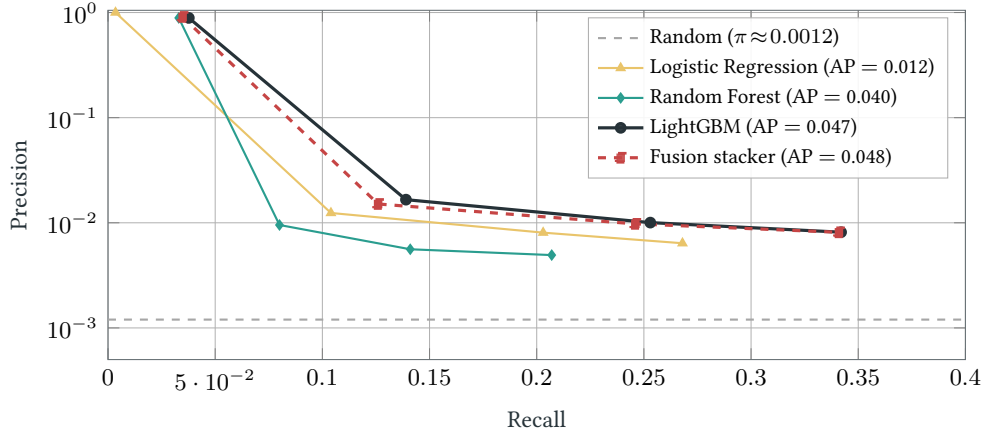

Figure~\ref{fig:prcurves} shows precision--recall curves reconstructed from the calibrated high-precision operating point (Table~\ref{tab:sup}) and the recall-at-top-$K\%$ anchors of Table~\ref{tab:topk} against the theoretical random baseline of $\pi \approx 0.0012$.

\begin{table}[t]
\centering
\caption{Supervised and fusion performance on the test set at the validation-calibrated $\sim 90\%$-precision operating point.}\label{tab:sup}
\small
\begin{tabular}{lccccc}
\toprule
\rowH \textbf{Model} & \textbf{PR-AUC} & \textbf{ROC-AUC} & \textbf{Precision} & \textbf{Recall} & \textbf{Alerts/10k} \\
\midrule
Logistic Regression & $0.0118$ & $0.7492$ & $1.0000$ & $0.0035$ & $0.04$ \\
Random Forest       & $0.0399$ & $0.6954$ & $0.8889$ & $0.0330$ & $0.44$ \\
LightGBM            & $\mathbf{0.0469}$ & $\mathbf{0.7897}$ & $0.8889$ & $\mathbf{0.0377}$ & $0.51$ \\
Fusion (stacker)    & $0.0477$ & $0.7856$ & $0.9077$ & $0.0348$ & $0.46$ \\
\bottomrule
\end{tabular}
\end{table}

\subsection{Confusion matrices at the calibrated threshold}
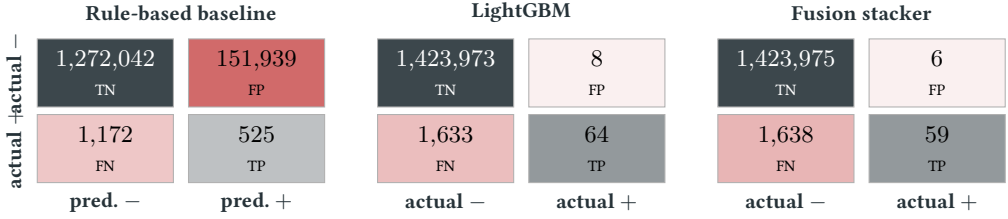
\begin{figure}[t]
\centering
\begin{tikzpicture}[
  font=\footnotesize,
  cell/.style={rectangle, draw=aimsrule, minimum width=18mm, minimum height=9mm, align=center},
  mylab/.style={font=\scriptsize\bfseries, color=aimsdark}
]
% Rule-based
\node[mylab] at (0, 3.0) {Rule-based baseline};
\node[cell, fill=aimsdark!90, text=white] at (-1, 2.2) {$1{,}272{,}042$\\ \tiny TN};
\node[cell, fill=aimsaccent!80] at ( 1, 2.2) {$151{,}939$\\ \tiny FP};
\node[cell, fill=aimsaccent!30] at (-1, 1.2) {$1{,}172$\\ \tiny FN};
\node[cell, fill=aimsdark!30] at ( 1, 1.2) {$525$\\ \tiny TP};
\node[mylab] at (-1, 0.5) {pred.\ $-$};
\node[mylab] at ( 1, 0.5) {pred.\ $+$};

% LightGBM
\node[mylab] at (4.5, 3.0) {LightGBM};
\node[cell, fill=aimsdark!90, text=white] at (3.5, 2.2) {$1{,}423{,}973$\\ \tiny TN};
\node[cell, fill=aimsaccent!8] at ( 5.5, 2.2) {$8$\\ \tiny FP};
\node[cell, fill=aimsaccent!35] at (3.5, 1.2) {$1{,}633$\\ \tiny FN};
\node[cell, fill=aimsdark!50] at ( 5.5, 1.2) {$64$\\ \tiny TP};
\node[mylab] at (3.5, 0.5) {actual~$-$};
\node[mylab] at ( 5.5, 0.5) {actual~$+$};

% Fusion
\node[mylab] at (9.0, 3.0) {Fusion stacker};
\node[cell, fill=aimsdark!90, text=white] at (8.0, 2.2) {$1{,}423{,}975$\\ \tiny TN};
\node[cell, fill=aimsaccent!8] at (10.0, 2.2) {$6$\\ \tiny FP};
\node[cell, fill=aimsaccent!40] at (8.0, 1.2) {$1{,}638$\\ \tiny FN};
\node[cell, fill=aimsdark!50] at (10.0, 1.2) {$59$\\ \tiny TP};
\node[mylab] at (8.0, 0.5) {actual~$-$};
\node[mylab] at (10.0, 0.5) {actual~$+$};

% Row labels on the left panel (rows are the actual class)
\node[mylab, rotate=90] at (-2.2, 2.2) {actual~$-$};
\node[mylab, rotate=90] at (-2.2, 1.2) {actual~$+$};
\end{tikzpicture}
\caption{Confusion-matrix comparison at the validation-calibrated high-precision operating point on the test window. The rule-based baseline (left) achieves higher raw recall ($525$ true positives) but at the cost of $\sim 152{,}000$ false alerts. LightGBM (centre) suppresses false positives to $8$ while capturing $64$ true laundering transactions. The fusion stacker (right) tightens further to $6$ false positives and recovers $59$ true positives, a small trade of both alert volume and captured cases relative to LightGBM. Fusion at this operating point acts as a stricter cut of the supervised score rather than as a corroborative recall booster.}
\label{fig:cm}
\end{figure}

Table~\ref{tab:cm} and Figure~\ref{fig:cm} make the absolute cost of the operating point explicit. On the $1.43$M-transaction test window, LightGBM at $\mathrm{Prec}\approx 0.89$ generates $72$ alerts of which $64$ are true positives, missing $1{,}633$ laundering transactions. The fusion stacker at $\mathrm{Prec}\approx 0.91$ generates $65$ alerts of which $59$ are true positives, missing $1{,}638$ laundering transactions: a small reduction in both alert volume and captured cases relative to LightGBM. A conservative rule-based baseline (single-transaction amount above a heavy-tail threshold, \emph{or} a cross-border transaction in which the sender and receiver bank locations differ) generates approximately $150{,}000$ alerts and captures $525$ laundering transactions, exhibiting the ``high-volume, low-precision'' failure mode characteristic of static rules in mobile money.

\begin{table}[t]
\centering
\caption{Confusion matrices at the calibrated high-precision operating point on the test window ($N = 1{,}425{,}678$ transactions; $\approx 1{,}697$ laundering positives). Rule-based baseline uses conventional amount-and-count thresholds.}\label{tab:cm}
\small
\begin{tabular}{lcccc}
\toprule
\rowH \textbf{Model} & \textbf{TN} & \textbf{FP} & \textbf{FN} & \textbf{TP} \\
\midrule
Rule-based baseline & $1{,}272{,}042$ & $151{,}939$ & $1{,}172$ & $525$ \\
LightGBM ($\mathrm{Prec}\approx 0.89$) & $1{,}423{,}973$ & $8$ & $1{,}633$ & $64$ \\
Fusion stacker ($\mathrm{Prec}\approx 0.91$) & $1{,}423{,}975$ & $6$ & $1{,}638$ & $59$ \\
\bottomrule
\end{tabular}
\end{table}

\subsection{Unsupervised and reconstruction detectors}
Table~\ref{tab:unsup} reports the anomaly and reconstruction detectors at a comparable alert budget of $\approx 1$ alert per $10{,}000$ transactions. The autoencoder is the strongest anomaly-only signal; Isolation Forest is uninformative at this budget; LOF is intermediate. All three are dominated by the supervised models at the same budget, consistent with the labelled-data regime SAML-D provides.

\begin{table}[t]
\centering
\caption{Unsupervised and reconstruction detector performance at $\approx 1$ alert per $10{,}000$ transactions on the test set.}\label{tab:unsup}
\small
\begin{tabular}{lccccc}
\toprule
\rowH \textbf{Model} & \textbf{PR-AUC} & \textbf{ROC-AUC} & \textbf{Precision} & \textbf{Recall} & \textbf{Alerts/10k} \\
\midrule
Isolation Forest & $0.0021$ & $0.6391$ & $0.0078$ & $0.0006$ & $0.90$ \\
Autoencoder      & $\mathbf{0.0210}$ & $0.6428$ & $\mathbf{0.2671}$ & $\mathbf{0.0253}$ & $1.13$ \\
Local Outlier Factor & $0.0099$ & $0.5685$ & $0.1558$ & $0.0141$ & $1.08$ \\
\bottomrule
\end{tabular}
\end{table}

\subsection{Capacity-aligned ranking: recall at top-$K\%$}
Table~\ref{tab:topk} reports recall when analysts work only the highest-scoring $K\%$ of the test window. LightGBM captures $13.9\%$ of laundering transactions in the top $1\%$ and $34.2\%$ in the top $5\%$; fusion is essentially indistinguishable at top $3$--$5\%$ (which is what one should expect: fusion's coefficients are dominated by the supervised score in the high-precision regime). The autoencoder and LOF contribute meaningfully only at deeper triage budgets.

\begin{table}[t]
\centering
\caption{Recall at top-$K\%$ highest-risk transactions on the test set. LightGBM and fusion are essentially co-ranked at $K = 3\%, 5\%$; fusion loses recall at $K = 1\%$.}\label{tab:topk}
\small
\begin{tabular}{lccc}
\toprule
\rowH \textbf{Model} & \textbf{Recall@$1\%$} & \textbf{Recall@$3\%$} & \textbf{Recall@$5\%$} \\
\midrule
Logistic Regression & $0.104$ & $0.203$ & $0.268$ \\
Random Forest       & $0.080$ & $0.141$ & $0.207$ \\
LightGBM            & $\mathbf{0.139}$ & $\mathbf{0.253}$ & $\mathbf{0.342}$ \\
Fusion              & $0.126$ & $0.246$ & $0.341$ \\
Autoencoder         & $0.061$ & $0.093$ & $0.114$ \\
LOF                 & $0.042$ & $0.089$ & $0.127$ \\
\bottomrule
\end{tabular}
\end{table}

\subsection{Reading the fusion result}\label{sec:fusiondiscussion}
Portfolio fusion is often presented as recall-enhancing. In this dataset, at the operating point relevant to Rwanda-style supervision, it is not. The stacker (Eq.~\ref{eq:fusion}) at $\mathrm{Prec}\approx 0.91$ recovers $59$ true positives against LightGBM's $64$ (a $\sim 8\%$ reduction), while lowering alert volume from $72$ to $65$ (a $\sim 10\%$ reduction) and raising precision by two percentage points. This is a small marginal trade rather than a genuine improvement: fusion is buying a small alert-volume saving at a small true-positive cost, without providing meaningfully new detection. The underlying reason is that the anomaly detectors are weakly informative on this dataset (Table~\ref{tab:unsup}), which pushes the fitted coefficients $\{\beta_2, \beta_3\}$ close to zero and effectively makes the stacker's high-precision operating point a stricter cut of the LightGBM score. Diagnostically, the fitted stacker assigns its largest absolute coefficient to the LightGBM score; the coefficients on the unsupervised and autoencoder inputs are near zero. This explains why fusion at the calibrated operating point behaves as a stricter cut of the LightGBM score, and why it produces neither a large recall gain nor a large alert-volume gain: the anomaly signals carry little information the supervised score does not already capture on this dataset. At maximally strict thresholds (as observed in supplementary experiments where the calibrated threshold approaches unity), the stacker degenerates further and issues only a handful of alerts. Fusion in this regime cannot compensate for the underlying anomaly-detector weakness.

The operational role of fusion in this deployment is therefore \emph{corroboration and audit-narrative}, not incremental recall: when an alert is flagged, presenting the analyst with concurring supervised and anomaly scores strengthens the case narrative supporting an STR. We do not recommend using the fusion stacker as the primary alert generator on this dataset; LightGBM at its calibrated high-precision operating point is the deployable ranking, with the anomaly scores presented alongside as complementary evidence.

\subsection{What drives the LightGBM ranking}
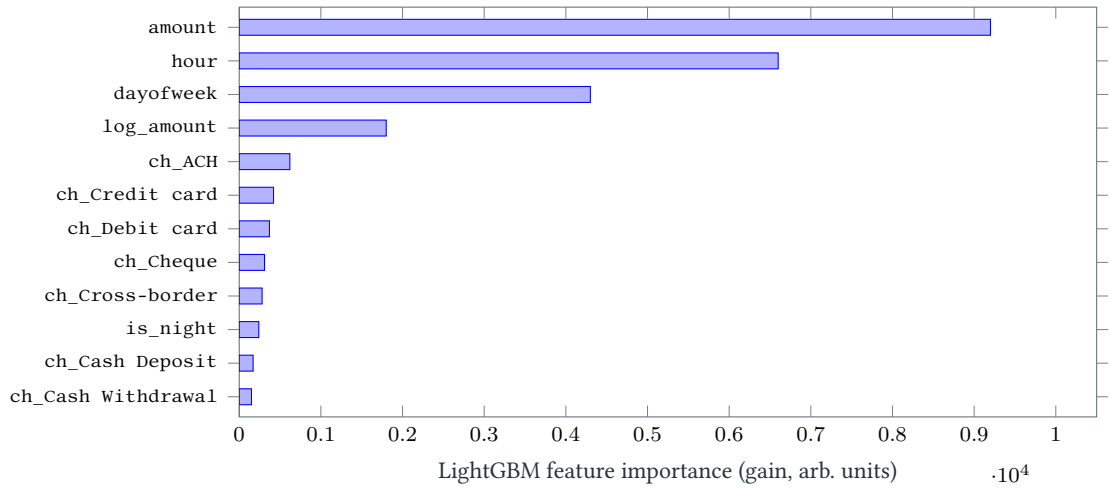
\begin{figure}[t]
\centering
\begin{tikzpicture}
\begin{axis}[
  width=0.85\linewidth, height=7cm,
  xbar,
  bar width=6pt,
  xlabel={LightGBM feature importance (gain, arb.\ units)},
  ytick={1,2,3,4,5,6,7,8,9,10,11,12},
  yticklabels={%
    \texttt{ch\_Cash Withdrawal},
    \texttt{ch\_Cash Deposit},
    \texttt{is\_night},
    \texttt{ch\_Cross-border},
    \texttt{ch\_Cheque},
    \texttt{ch\_Debit card},
    \texttt{ch\_Credit card},
    \texttt{ch\_ACH},
    \texttt{log\_amount},
    \texttt{dayofweek},
    \texttt{hour},
    \texttt{amount}%
  },
  xmin=0, xmax=10500,
  enlarge y limits={abs=0.6},
  axis line style={aimsdark!70},
  tick label style={font=\scriptsize},
  label style={font=\footnotesize\color{aimsdark}},
  every axis plot/.append style={fill=aimsdark!70, draw=aimsdark}
]
\addplot coordinates {
  (150,1) (170,2) (240,3) (280,4)
  (310,5) (370,6) (420,7) (620,8)
  (1800,9) (4300,10) (6600,11) (9200,12)
};
\end{axis}
\end{tikzpicture}
\caption{Top-12 features by LightGBM gain-based importance on the training set (values reproduced from the corresponding thesis experiment). Amount, hour, and day-of-week dominate the ranking, followed by log-amount and channel indicators; per-alert SHAP explanations should be delivered to analysts to translate global importance into per-instance narratives.}
\label{fig:featimp}
\end{figure}

Top-ranked drivers of the LightGBM score by gain-based importance are shown in Figure~\ref{fig:featimp}. Transaction amount, hour-of-day, day-of-week, and $\log$-transformed amount dominate the top of the ranking; channel indicators (with cross-border and cash-conversion channels most influential) follow. Time-aware aggregates dominate over static amount features once behavioural windows are included, consistent with the finding that structuring and rapid-layering behaviours are best detected in the temporal derivative of activity rather than in raw amounts.

\subsection{Where the model still fails}
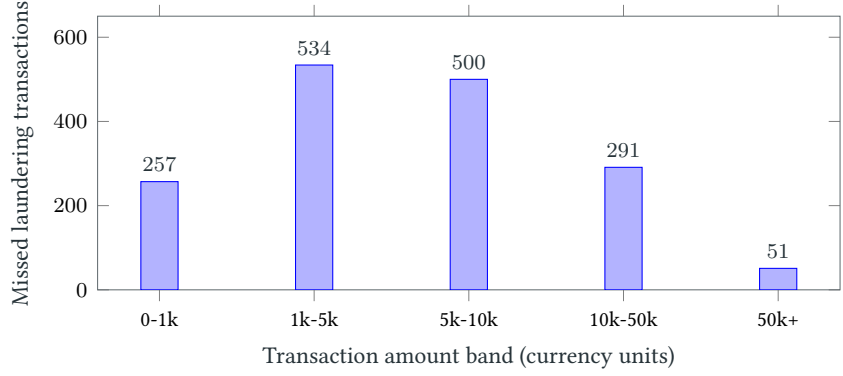
\begin{figure}[t]
\centering
\begin{tikzpicture}
\begin{axis}[
  width=0.75\linewidth, height=5.2cm,
  ybar,
  bar width=14pt,
  xlabel={Transaction amount band (currency units)},
  ylabel={Missed laundering transactions},
  symbolic x coords={0-1k, 1k-5k, 5k-10k, 10k-50k, 50k+},
  xtick=data,
  ymin=0, ymax=650,
  nodes near coords,
  nodes near coords style={font=\scriptsize, color=aimsdark},
  axis line style={aimsdark!70},
  tick label style={font=\scriptsize},
  label style={font=\footnotesize\color{aimsdark}},
  every axis plot/.append style={fill=aimsaccent!70, draw=aimsaccent}
]
\addplot coordinates {
  (0-1k,257)
  (1k-5k,534)
  (5k-10k,500)
  (10k-50k,291)
  (50k+,51)
};
\end{axis}
\end{tikzpicture}
\caption{Distribution of false negatives (missed laundering transactions) by amount band at the LightGBM calibrated high-precision operating point on the test window. Failures concentrate in the mid-band ($1{,}000$--$10{,}000$ currency units): transactions that are neither anomalous in amount nor rare in tempo, and therefore blend into legitimate merchant and remittance activity. High-value cases ($\geq 50{,}000$) and sub-$1{,}000$ structuring are captured comparatively well.}
\label{fig:fnamount}
\end{figure}

Table~\ref{tab:fn} and Figure~\ref{fig:fnamount} distribute missed laundering transactions by amount band. False negatives concentrate in the mid-band ($1{,}000$--$10{,}000$ currency units), where a single transaction is neither anomalous in amount nor rare in tempo. High-value laundering ($>50{,}000$) is captured reliably; sub-$1{,}000$ structuring is caught primarily via the rolling-count features. Read against the Rwanda-relevant typologies of Section~\ref{sec:regcontext}, the mid-band failure mode corresponds most directly to (b) fan-in patterns onto merchant or business wallets and (c) agent-mediated cash conversion, both of which blend into legitimate high-frequency retail activity at sub-hour scales. It is the natural target for a graph-based enrichment layer in future work.

\begin{table}[t]
\centering
\caption{False negatives by amount band on the test set at the LightGBM high-precision operating point.}\label{tab:fn}
\small
\begin{tabular}{lc}
\toprule
\rowH \textbf{Amount band} & \textbf{False negatives (count)} \\
\midrule
$[0, 1{,}000)$        & $257$ \\
$[1{,}000, 5{,}000)$  & $\mathbf{534}$ \\
$[5{,}000, 10{,}000)$ & $\mathbf{500}$ \\
$[10{,}000, 50{,}000)$& $291$ \\
$\geq 50{,}000$       & $51$ \\
\bottomrule
\end{tabular}
\end{table}

%======================================================================
\section{Ablation and sensitivity}\label{sec:ablation}

Table~\ref{tab:abl} reports LightGBM performance under three feature-group ablations and one training ablation, on the test window. Removing behavioural features (rolling aggregates, velocity, counterparty diversity, net flows) causes the largest drop in PR-AUC ($-33\%$). Operational lift comes from temporal-behavioural structure rather than from raw amounts or channel metadata. Removing time-context features (hour, day-of-week, night flag) costs $-16\%$; removing cost-sensitive class weights and training on the imbalanced distribution costs $-39\%$ recall at the high-precision operating point.

\begin{table}[t]
\centering
\caption{Ablation on LightGBM: PR-AUC and recall at the calibrated $\sim 90\%$-precision operating point, test set.}\label{tab:abl}
\small
\begin{tabular}{lcc}
\toprule
\rowH \textbf{Configuration} & \textbf{PR-AUC} & \textbf{Recall @ Prec $\approx 0.9$} \\
\midrule
Full model                       & $0.0469$ & $0.0377$ \\
$\;-$ behavioural features       & $0.0315$ & $0.0234$ \\
$\;-$ time-context features      & $0.0392$ & $0.0291$ \\
$\;-$ class weighting            & $0.0287$ & $0.0186$ \\
\bottomrule
\end{tabular}
\end{table}

\subsection*{Temporal stability check (short-horizon)}
Splitting the test window into three equal-duration sub-windows and evaluating LightGBM independently on each yields PR-AUC values of $0.048$, $0.046$, and $0.045$. Ranking quality is preserved across the short horizon SAML-D spans ($\sim 10$ months of simulated time); this is \emph{not} evidence of robustness to multi-year drift (see Section~\ref{sec:limits}).

%======================================================================
\section{Operational mapping for Rwanda-style mobile money}\label{sec:opmapping}

\subsection{Score-to-band mapping and analyst workflow}
Let $s(x)$ denote the deployed LightGBM score. Using the validation window to fix thresholds $\tau_1 < \tau_2 < \tau_3$ against workload and precision targets, we define four action bands (Table~\ref{tab:bands}).

\begin{table}[t]
\centering
\caption{Risk bands, thresholds, and analyst actions in the proposed Rwanda-aligned deployment. Thresholds are illustrative and would be recalibrated on real deployment data.}\label{tab:bands}
\small
\begin{tabular}{p{2.2cm}p{2.6cm}p{3.6cm}p{5cm}}
\toprule
\rowH \textbf{Band} & \textbf{Score range} & \textbf{Alert budget / $10^4$} & \textbf{Action} \\
\midrule
A (monitor) & $s(x) < \tau_1 = 0.10$ & 0 & Passive monitoring; no case creation.\\
B (triage)  & $\tau_1 \leq s(x) < \tau_2 = 0.60$ & $\sim 20$ & Lightweight analyst check: history, velocity flags, counterparty profile.\\
C (priority review) & $\tau_2 \leq s(x) < \tau_3 = 0.95$ & $\sim 3$ & Case creation, enhanced due diligence, cross-account link checks. \\
D (critical) & $s(x) \geq \tau_3 = 0.95$ & $\sim 0.5$ & Fast-track escalation for STR/SAR decisioning subject to policy and legal procedures.\\
\bottomrule
\end{tabular}
\end{table}

The high-precision operating point reported in Table~\ref{tab:sup} corresponds to Band~D. Band~C is designed as the working queue for analysts on a routine daily cadence; Band~B supports batched review and monthly audit sampling.

\subsection{Explanation payload delivered with each alert}
Each alert dispatched to an analyst carries a compact behavioural explanation: (i)~the top-three feature contributions from a local SHAP explanation of the LightGBM score \citep{lundberg2017shap}; (ii)~deviation of the account from its own baseline on the same features over a $30$-day trailing window; (iii)~corroboration indicators from the anomaly detectors (autoencoder reconstruction score above the $99.5$th validation percentile; Isolation Forest score above threshold). This payload is designed for direct inclusion in the STR narrative under FIC's reporting template.

\subsection{Operational impact: from alerts per $10{,}000$ to analyst hours}\label{sec:opimpact}
The abstract detection metrics translate directly into investigator workload once a monthly transaction volume is assumed. Table~\ref{tab:opimpact} illustrates the calculation at three volume scenarios plausible for the Rwandan retail mobile-money segment, applied to (a) the rule-based baseline, (b) LightGBM at its calibrated high-precision operating point, and (c) the fusion stacker. Under any of these scenarios, the rule-based baseline generates alert volumes an order of magnitude beyond feasible investigator capacity, while the calibrated ML pipeline produces a workload that fits within a small dedicated team.

\begin{table}[t]
\centering
\caption{Illustrative operational load at three hypothetical daily transaction volumes for the Rwandan retail mobile-money segment. Assumptions (all clearly hypothetical): $20$~minutes analyst-time per alert triage, $6$ productive analyst-hours per shift, all alerts triaged same day. Actual BNR/provider values must replace these before any deployment claim.}\label{tab:opimpact}
\small
\begin{tabular}{lrrrr}
\toprule
\rowH \textbf{Detector} & \textbf{Alerts/10k} & \textbf{Daily alerts} & \textbf{Analyst-hours/day} & \textbf{FTE analysts} \\
\midrule
\multicolumn{5}{l}{\textit{Scenario A: $1$M transactions/day}} \\
Rule-based baseline & $\sim 1{,}066$ & $106{,}600$ & $35{,}533$ & infeasible \\
LightGBM (Prec$\approx 0.89$) & $0.51$ & $51$ & $17.0$ & $2.8$ \\
Fusion stacker & $0.46$ & $46$ & $15.3$ & $2.6$ \\
\midrule
\multicolumn{5}{l}{\textit{Scenario B: $5$M transactions/day}} \\
Rule-based baseline & $\sim 1{,}066$ & $533{,}000$ & $177{,}666$ & infeasible \\
LightGBM (Prec$\approx 0.89$) & $0.51$ & $255$ & $85$ & $14.2$ \\
Fusion stacker & $0.46$ & $230$ & $77$ & $12.8$ \\
\midrule
\multicolumn{5}{l}{\textit{Scenario C: $10$M transactions/day}} \\
Rule-based baseline & $\sim 1{,}066$ & $1{,}066{,}000$ & $355{,}333$ & infeasible \\
LightGBM (Prec$\approx 0.89$) & $0.51$ & $510$ & $170$ & $28.3$ \\
Fusion stacker & $0.46$ & $460$ & $153$ & $25.6$ \\
\bottomrule
\end{tabular}
\end{table}

The pattern is stable across scenarios: the calibrated ML pipeline reduces investigator workload by roughly three orders of magnitude relative to a conventional rule-based baseline at the same precision expectations. At Scenario~B (a plausible mid-range volume for the combined MTN Rwanda and Airtel Rwanda footprint), LightGBM's $\sim 255$ daily alerts correspond to a full-time compliance team of $12$--$14$ analysts, which is within the range of a supervised financial institution's dedicated AML function. In the same scenario the rule-based baseline is not operable: no plausible team size can process $\sim 533{,}000$ daily alerts, forcing analysts into arbitrary sampling and therefore into structurally biased case selection.

The caveat is that recall at this workload is low in absolute terms ($\approx 3.8\%$ at the high-precision operating point). What the ML pipeline offers is not ``finding most laundering'' but surfacing the small subset of alerts most likely to be laundering. The remainder of the true-positive base is recovered through complementary channels: STR filings from external tips, cross-institution intelligence sharing, and audit-driven review of top-$K\%$ ranked activity. Framing the tool this way matters for setting realistic expectations with supervisors.

\subsection{Institutional workflow}
\begin{figure}[t]
\centering
\begin{tikzpicture}[
  font=\small,
  node distance=6mm and 8mm,
  every node/.style={align=center},
  stage/.style={rectangle, rounded corners=2pt, draw=pnavy, thick, minimum width=28mm, minimum height=11mm, fill=pnavy!8},
  band/.style={rectangle, rounded corners=2pt, draw=pochre, minimum width=22mm, minimum height=8mm, fill=pochre!18},
  inst/.style={rectangle, rounded corners=2pt, draw=pmoss, thick, minimum width=25mm, minimum height=10mm, fill=pmoss!14},
  arr/.style={-Latex, thick},
  darr/.style={-Latex, dashed}
]
% Stage 1
\node[stage] (capture) {Transaction capture\\ \footnotesize (MTN MoMo, Airtel, banks)};
\node[stage, right=of capture] (score) {ML scoring\\ \footnotesize (LightGBM + anomaly)};
\node[stage, right=of score] (band) {Risk banding\\ \footnotesize (A / B / C / D)};

% Bands
\node[band, below=8mm of band, xshift=-32mm] (bandA) {Band A\\ \footnotesize monitor};
\node[band, right=4mm of bandA] (bandB) {Band B\\ \footnotesize triage};
\node[band, right=4mm of bandB] (bandC) {Band C\\ \footnotesize priority};
\node[band, right=4mm of bandC] (bandD) {Band D\\ \footnotesize critical};

% Analyst
\node[stage, below=14mm of bandB, xshift=15mm] (analyst) {Analyst review\\ \footnotesize (SHAP payload,\\ baseline deviation)};

% Decisions
\node[stage, below=8mm of analyst, xshift=-25mm] (close) {Close\\ \footnotesize (false alert)};
\node[stage, below=8mm of analyst] (edd) {Enhanced\\due diligence};
\node[stage, below=8mm of analyst, xshift=25mm] (escalate) {Escalate\\ \footnotesize (STR case)};

% Institutions
\node[inst, below=10mm of escalate, xshift=-15mm] (fic) {\textbf{FIC}\\ \footnotesize STR/SAR};
\node[inst, right=8mm of fic] (bnr) {\textbf{BNR}\\ \footnotesize supervision};

% Arrows — top row
\draw[arr] (capture) -- (score);
\draw[arr] (score) -- (band);
\draw[arr] (band) -- (bandA);
\draw[arr] (band) -- (bandB);
\draw[arr] (band) -- (bandC);
\draw[arr] (band) -- (bandD);

% Bands to analyst
\draw[arr] (bandB) -- (analyst);
\draw[arr] (bandC) -- (analyst);
\draw[arr] (bandD) -- (analyst);

% Analyst decisions
\draw[arr] (analyst) -- (close);
\draw[arr] (analyst) -- (edd);
\draw[arr] (analyst) -- (escalate);

% Escalation to institutions
\draw[arr] (escalate) -- (fic);
\draw[darr] (fic) -- (bnr) node[midway,above,font=\scriptsize]{coordinate};

\end{tikzpicture}
\caption{Rwanda-aligned operational workflow: transaction capture at mobile-money providers, ML scoring, risk banding (Bands A--D as defined in Table~\ref{tab:bands}), analyst review with SHAP-based explanation payload, and escalation to the Financial Intelligence Centre (FIC) with supervisory coordination through the National Bank of Rwanda (BNR).}
\label{fig:workflow}
\end{figure}
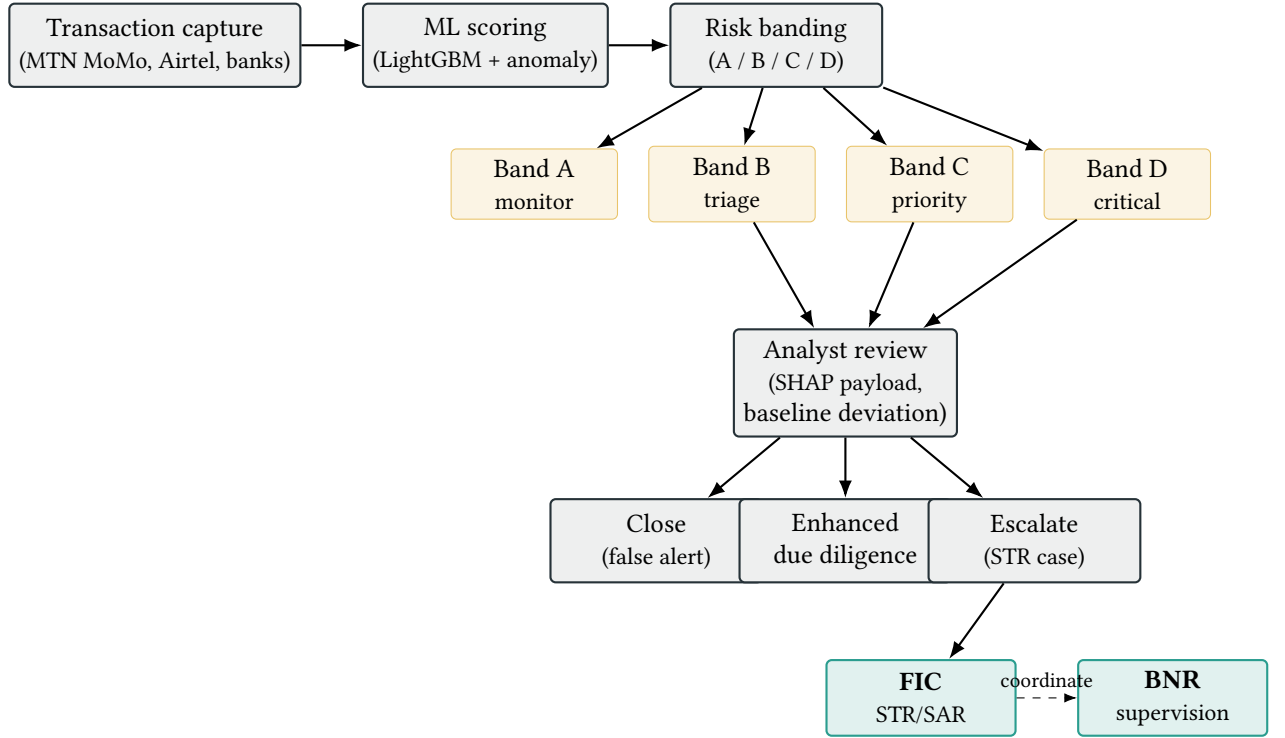

Alerts flow through four operational stages illustrated in Figure~\ref{fig:workflow}: (1)~transaction capture and scoring at the provider (MTN Mobile Money, Airtel Money, or bank-led wallet); (2)~internal analyst triage using the band schema above; (3)~compliance review and case escalation; (4)~STR/SAR filing to FIC. Model documentation (feature specifications, hyperparameter versions, threshold-change approvals, and monitoring-report cadence) is maintained as a compliance artefact aligned to ESAAMLG recommendations \citep{esaamlg2023}.

\subsection{Ethics, privacy, and governance}
An AML monitoring system generates real customer harm when misused: unnecessary account restrictions, biased targeting of vulnerable segments, opaque decisioning. Any deployment on real Rwandan mobile-money data must enforce (i)~purpose limitation (scores support investigation, they do not automate account restriction); (ii)~access control and audit logs for sensitive customer data; (iii)~human-in-the-loop review for all adverse actions; (iv)~periodic stratified performance and error-rate checks across KYC tiers, customer segments (rural/urban, individual/merchant), and channels. In the current work we do not have access to protected-attribute proxies in SAML-D and therefore do not report group-wise metrics; this is a real limitation, addressed in Section~\ref{sec:limits}.

%======================================================================
\section{Limitations, threats to validity, and deployment path}\label{sec:limits}

\subsection{Synthetic data}
SAML-D enables reproducible benchmarking but cannot capture the full distribution of operational noise, label uncertainty, and adversarial adaptation of real transaction streams. It also abstracts several mobile-money-specific attributes: KYC tiers, device and SIM identifiers, agent identifiers and topology, and network structure across accounts. Results should be interpreted as evidence for a pipeline and evaluation protocol, not for a production-calibrated national risk model.

\subsection{Drift claims}
The SAML-D simulation spans approximately ten months of simulated time. The temporal-stability check in Section~\ref{sec:ablation} therefore evidences \emph{short-horizon} rank stability only. Multi-year drift (driven by product changes, regulatory shifts, macroeconomic cycles, or adversarial adaptation) is not observable in this dataset, and no claim about robustness to such drift is made here. Any real deployment would require rolling-window backtesting and scheduled recalibration; the pipeline is designed to support this, but the empirical evidence is not.

\subsection{Fairness}
We do not report group-wise recall, precision, or alert-rate disparities because SAML-D does not encode protected-attribute proxies (KYC tier, geographic region, urban/rural, individual/merchant) in a way that reflects the Rwandan population. This is a real gap. A minimum standard for real-data deployment is stratified evaluation across at least KYC tier, geographic zone, and customer type, with disparate-impact and equalised-odds monitoring on a scheduled cadence.

\subsection{Fusion result generality}
The fusion analysis in Section~\ref{sec:fusiondiscussion} is specific to SAML-D. Data with richer network structure or with anomaly signals that carry more independent information than they do here may yield a different balance; the conclusion drawn from this evidence does not generalise beyond it.

\subsection{Uncertainty on reported metrics}
Metrics reported in Section~\ref{sec:results} are point estimates from a single chronological split with a single training seed. Two sources of uncertainty are not quantified here and are deferred to a follow-up. First, \emph{test-set variance}: a paired bootstrap over the test-set predictions at the fixed calibrated thresholds ($B = 1{,}000$ resamples with replacement, computing PR-AUC, high-precision recall, alerts per $10{,}000$, and confusion-matrix counts on each resample) would yield $95\%$ confidence intervals on each headline metric; the paired-bootstrap protocol is implemented in the companion notebook and can be executed on the trained models directly. Second, \emph{seed variance}: training-time randomness (LightGBM's histogram binning, autoencoder initialisation and mini-batch order, Random Forest's bootstrap sampling) contributes additional variability not visible from a single run; the minimum protocol is to repeat training across $B \geq 10$ seeds and report the median and $[5\%,\,95\%]$ range of each headline metric. Given that the emphasis of this proof-of-concept study is the pipeline and the operational calibration protocol rather than point-estimate optimisation, we treat the reported values in Section~\ref{sec:results} as evidence for the pipeline and evaluation protocol rather than as production-calibrated bounds; any journal or supervisory submission of a deployed model should include both interval reports above.

\subsection{Comparison to prior baselines on SAML-D}
Our high-precision operating results are broadly consistent with the LightGBM-tier numbers reported by \citet{oztas2024saml} on the same data. We do not claim algorithmic novelty over that baseline. The contribution of this work relative to it is the calibrated operating regime, the confusion-matrix analysis of the fusion result, the recall-at-top-$K\%$ triage frame, and the operational mapping to Rwandan institutions.

\subsection{From synthetic to real deployment}
A viable transition path proceeds in five stages. (1)~\emph{Secure onboarding}: a privacy-compliant feature store built on causal aggregation, hosted under BNR-approved data-handling controls, with encrypted linkage to provider transaction logs. (2)~\emph{Shadow mode}: parallel scoring alongside the existing rule-based system for a bounded period (three to six months), measuring incremental true-positive lift and alert-rate impact without changing operational decisions. (3)~\emph{Calibration with real outcomes and stratified fairness evaluation}: threshold recalibration using confirmed STR/SAR outcomes and analyst feedback, together with group-wise performance monitoring across KYC tier, geographic zone, and customer type (individual/merchant), reported to compliance on a scheduled cadence. (4)~\emph{Incremental enrichment}: staged addition of graph-based features once account-level linkage across providers is legally and technically feasible, and of sequence models once labelled sequences accumulate. (5)~\emph{Independent validation}: periodic third-party model validation, model-card documentation, and supervisory reporting integrated with FIC and BNR compliance change-management.

%======================================================================
\section{Conclusion}\label{sec:conclusion}

We presented a machine-learning framework for AML transaction monitoring in Rwanda's mobile-money ecosystem, developed against explicit operational constraints: extreme class imbalance, capacity-bounded investigation, delayed labels, and continuous drift. On SAML-D, LightGBM with causal behavioural features attains the strongest calibrated single-model ranking (PR-AUC $0.0469$; $64$ true positives at precision $\approx 0.89$ on the test window), and a logistic-regression late-fusion stacker delivers a small alert-volume reduction ($72\!\to\!65$ alerts) at a small true-positive cost ($64\!\to\!59$) without adding recall. We reported this fusion result as it is: corroboration and audit-narrative rather than incremental recall. We mapped supervised scores to a four-band analyst workflow aligned to the Financial Intelligence Centre's reporting regime and to ESAAMLG evaluation guidance, and made explicit the translation from alerts per $10{,}000$ to analyst-hours under three transaction-volume scenarios. The contribution is operational: a governance-aware pipeline and evaluation protocol calibrated for an African mobile-money regulator, together with a staged deployment path from synthetic prototyping to real-data validation under BNR and FIC oversight. The pipeline transfers directly to real mobile-money platforms; the numbers do not, and should be recalibrated on real data before any deployment decision.

\section*{Declaration of competing interest}
The authors declare that they have no known competing financial interests or personal relationships that could have appeared to influence the work reported in this paper.

\section*{Data and code availability}
The SAML-D dataset is publicly available on Kaggle \citep{oztaskaggle2023}. Feature-engineering, model-training, and evaluation code will be made available on request pending institutional review.

\section*{Acknowledgements}
The authors thank the African Institute for Mathematical Sciences (AIMS) Senegal for computational resources and the tutor and academic staff for scientific guidance. No real customer or transaction data were used at any stage of this study.

\bibliography{refs}

\end{document}